\documentclass{svproc}
\usepackage{url}

\usepackage{subcaption}
\usepackage{xcolor}
\usepackage{amsmath}
\usepackage{float}
\usepackage{siunitx} 
\usepackage{sidecap}
\usepackage{graphicx}
\usepackage{sidecap}
\usepackage{cancel}
\usepackage{commath}
\usepackage{amssymb}
\usepackage{mathtools}
\usepackage{graphicx}
\usepackage{physics}
\usepackage[utf8]{inputenc}
\usepackage[T1]{fontenc}
\begin{document}
\mainmatter     
\title{Long Short-Term Memory Neural Network for Temperature Prediction in Laser Powder Bed Additive Manufacturing}
\titlerunning{Temperature Prediction in Additive Manufacturing using LSTM}
%-----------------------------------------------------------------------------
%-----------------------------------------------------------------------------
%-----------------------------------------------------------------------------
% --------------- Please add Acknowledgment at the end -----------------------
%-----------------------------------------------------------------------------
%-----------------------------------------------------------------------------
%-----------------------------------------------------------------------------
%\author{}
%\authorrunning{} % abbreviated author list (for running head)
%\tocauthor{}
%\institute{}
%-----------------------------------------------------------------------------
%-----------------------------------------------------------------------------
%-----------------------------------------------------------------------------
%-----------------------------------------------------------------------------
%-----------------------------------------------------------------------------
%-----------------------------------------------------------------------------
%-----------------------------------------------------------------------------
\author{Ashkan Mansouri Yarahmadi \and Michael Breu\ss{} \and  Carsten Hartmann}
\authorrunning{Mansouri Yarahmadi et al.} % abbreviated author list (for running head)
\institute{
           BTU Cottbus-Senftenberg \\
           Institute for Mathematics \\
           Platz der Deutschen Einheit 1, \\ 
           03046 Cottbus, Germany \\
           \email{yarahmadi,breuss,hartmanc}@b-tu.de}
%-----------------------------------------------------------------------------
\maketitle  
\begin{abstract}
In context of laser powder bed fusion (L-PBF), it is known that the properties of the 
final fabricated product highly depend on the temperature distribution and its 
gradient over the manufacturing plate. 
In this paper, we propose a novel means to predict the temperature gradient distributions during the printing process by making use of neural networks. 
This is realized by employing heat maps produced by an optimized printing protocol simulation and used for training a specifically tailored recurrent neural network in terms of a long short-term memory architecture. 
The aim of this is to avoid extreme and inhomogeneous temperature distribution that may occur across the plate in the course of the printing process.  

In order to train the neural network, we adopt a well-engineered simulation and unsupervised learning framework. To maintain a minimized average thermal gradient across the plate, a cost function is introduced as the core criteria, which is inspired and optimized by considering the well-known traveling salesman problem (TSP). As time evolves the unsupervised printing 
process governed by TSP produces a history of temperature heat maps that maintain minimized average thermal gradient. 

All in one, we propose an intelligent printing tool that provides control over the substantial printing process components for L-PBF, i.e.\ optimal nozzle trajectory deployment 
as well as online temperature prediction for controlling 
printing quality. 
\end{abstract}
\section{Introduction}
In contrast to traditional machining, additive manufacturing (AM) builds objects layer by layer through a joining process of materials making the fabrication of individualized components possible across different engineering fields. The laser powder bed fusion (L-PBF) technique as an AM process, that we focus on in this study, uses a deposited powder bed which is selectivity fused by a computer-controlled laser beam~\cite{SHRUBTWZ2020}.
% neu:
The extreme heating by the laser on the one hand, and on the other hand
the influence of the degree of homogeneity of the heat distribution on 
the printing quality in L-PBF, make it highly challenging to conduct the 
printing process in an intelligent way that may guarantee high quality 
printing results.
As explained in more detail when discussing related work, there has thus
been a continuous effort to {\em (i)} propose beneficial printing paths that help to avoid unbalanced heating and
{\em (ii)} to forecast the heat distribution in order assess the potential
printing quality and terminate printing in case of foreseeable flaws.

In this paper, we propose to couple a laser beam trajectory devised on the basis of a heuristic control during the fabrication phase of L-PBF with prediction based on neural networks. 
The developed novel framework addresses both 
the abovementioned main issues in L-PBF and represents
an intelligent printing tool that provides control over the printing process. 
To this end, we aim at conducting controlled laser beam simulation that approximately achieves \textit{temperature constancy} on a simulated melted power bed. In addition, we opt to perform temperature rate of change prediction as an important factor for microscopic structure of the final fabricated product. 

The main novelty of the current paper is to adopt~\textit{long-short-term memory} (LSTM)~\cite{HS1997} prediction framework, which is introduced in Section~\ref{sec:LSTM} to predict the temperature distribution and its gradient during printing. This consequently can be used to avoid any overheating by taking necessary actions in advance, namely stopping the printing process to avoid the printer damage due to overheated deformed parts of the printing product. 
Based on this, we conjecture that our developed pipeline may provide a highly valuable step for practical printing that provides quality control of the printed product, while being efficient with regard to energy consumption and use of material. Finally, in Section~\ref{sec:Results}, we present an effective numerical test concerning the predicted temperature gradients.

In Section~\ref{sec:set-up} of this paper a simulation framework is brought out by recalling the heat transfer model together with a cost function that consists of two terms aiming to maintain almost a constant temperature with a low spatial gradient across the power bed area. For simplicity, we confine ourselves to a 2-dimensional domain, which is still a realistic description of printing over the manufacturing plate. In Subsection~\ref{sec:TSP}, the idea of the travelling salesman problem (TSP) as a heuristics for the laser beam steering is explained; being one of the most fundamental and well-studied NP-hard problems in the field of combinatorial optimization (e.g.~\cite{F1956,GJ1990}), we will use a stochastic optimization strategy (simulated annealing) to establish an optimal laser trajectory. 

%%%%%%%%%%%%%%%%%%%%%%%%%%%%%%%%%%%%%%%%%%%%%%%%%
\section{Related work in Laser Powder Bed Additive Manufacturing}
\label{sec:set-up}
%%%%%%%%%%%%%%%%%%%%%%%%%%%%%%%%%%%%%%%%%%%%%%%%%

In general, a variety of different laser beam parameters such as laser power, scan speed, building direction and laser thickness influence the final properties of the fabricated product. Due to intensive power of laser during additive manufacturing, the printed product can have defects, such as deviations from the target geometry or cracks caused by large temperature gradients. For example, inhomogeneous heating may lead to unmelted powder particles that can locally induce pores and microscopic cracks \cite{G-et-al-2021}. At the same time, the cooling process determines the microstructure of the printed workpiece and thus its material properties, such as strength or toughness, which depend on the proportion of carbon embedded in the crystal structure of the material \cite{A-et-al-2019}. 

%% noch ne Überleitung...?
In a broader view, machine learning approaches may be deployed to provide monitoring capabilities over varying factors of L-PBF, namely the used metal powder and its properties both at the initial time of spread and during the printing process as well as the laser beam parameters, aiming to investigate and avoid any defect generation during the fabrication. See~\cite{ARNST2017} for an survey.

Concerning the powder properties, different capturing technologies along with machine learning tools are used to automate the task of defect detection and avoidance during the printing process. In~\cite{SB2018,SB2018-CNN}, the k-means clustering~\cite{Matlab-KNN} and convolution neural network (CNN)~\cite{LB2015} respectively, were used to detect and classify defects at the time of initial powder spread and their probable consequences during the entire printing phase and based on captured grey images. In~\cite{KZZ2014}, high resolution temporal and evolving patterns are captured using a commercial EOS M270 system to find layer-wise heat inhomogeneities induced by the laser. In~\cite{KSF2015}, an inline coherent imaging (ICI) system was used to monitor the defects and unstable process regimes concerning the morphology changes and also the stability of the melt pools. Here, the back scattered intensities from the melt pool samples are measured as a function of their heights called A-lines. Later, a Gaussian fitting of individual A-lines is performed to determine centroid height and amplitude of melt pools as a function of time corresponding to a range of different stainless steel powders with different properties.

About the laser beam and its parameter optimization task one can avoid conducting expensive real experiments, in terms of material and power usage, by simulating the printing process by means of finite element method~\cite{FB2007} (FEM), Lattice Boltzmann method (LBM) or finite volume method (FVM) See~\cite{BSM2018,SCS2017} for extensive surveys. Later the gathered simulated data may be used in a data-driven machine learning approach within a L-PBF framework. In this context, a prediction task of thermal history was performed in~\cite{M-et-al-2018} by adopting a recurrent neural network (RNN) structure with a Gated Recurrent Unit (GRU) in a L-PBF process. A range of different geometries are simulated by FEM while accounting for different laser movement strategies, laser power and scan speed. A three-dimensional FEM is adopted in~\cite{ZC2006} to simulate the laser beam trajectory and investigate its effects on the residual stresses of the parts. The simulation results show modifications of the residual stress distributions and their magnitudes, that was validated through experimental tests, as a result of varying laser beam trajectory type. A parametric study~\cite{ZC2008-parametric} used the same FEM simulation setup as~\cite{ZC2006} with three varying factors namely the laser beam speed, the layer thickness and the laser deposition path width. While each factor value varies in its range from low, medium to high the hidden relations among the factors and their affects on residual stresses and part distortions are revealed.
 
In context of FEM simulation with a steering source of heat to represent the laser movement, one can refer to the work developed in~\cite{SFZSLMWYORK2020}. Here, the residual stresses during the printing is predicted though the laser nozzle steering rule is not revealed.

%%%%%%%%%%%%%%%%%%%%%%%%%%%%%%%%%%%%%%%%%%%%%%%%%
\section{Heat transfer model and TSP formulation}
\label{sec:set-up}
%%%%%%%%%%%%%%%%%%%%%%%%%%%%%%%%%%%%%%%%%%%%%%%%%
As indicated, we first describe our heat simulation setting which is the framework for the TSP optimization protocol described in the second part of this section.
%%%%%%%%%%%%%%%%%%%%%%%%%%%%%%%%%%%%%%
\subsection{Heat simulation framework}
%%%%%%%%%%%%%%%%%%%%%%%%%%%%%%%%%%%%%%
We set up a simulation environment, namely \textit{(i)} a moving source of heat (cf.~\eqref{eq:Heat}) to act as a laser beam on \textit{(ii)} an area $\Omega\subset\mathbb{R}^2$ simulated as deposition of aluminium metal powder called a plate. We assume that the plate is mounted to a base plate with large thermal conductivity, which makes the choice of Dirichlet boundary conditions with constant boundary temperature appropriate; if the surrounding is an insulator, then a reflecting i.e. zero-flux or von Neuman boundary condition is more suitable. A sequence of laser beam movements, called a trajectory, is followed so that at each point the heat equation~\eqref{eq:HE} is resolved based on FEM providing us a temperature map that varies on different plate locations as the time evolves.

Letting $u$ be the temperature across an open subset $\Omega\subset\mathbb{R}^2$ as time $t$ evolves in $\in[0,T]$, the heat equation that governs the time evolution of $u$ reads
\begin{subequations}\label{eq:HE}
\begin{align}\label{eq:HE1}
      \frac{\partial}{\partial t}u(x,y,t) & = \alpha \nabla^2 u(x,y,t) + \beta I(x,y)\,, & & (x,y,t)\in \Omega^\circ\times(0,T)\\\label{eq:HE2}
      u(x,y,t) & = \theta_0 & & (x,y,t)\in \partial\Omega\times[0,T] \\\label{eq:HE3}
      u(x,y,0) & = u_0(x,y) & & (x,y)\in \Omega
\end{align}
\end{subequations}
where we denote by 
\begin{equation}
\nabla^2 \phi= \frac{\partial^2 \phi}{\partial x^2} + \frac{\partial^2 \phi}{\partial y^2}
\label{eq:Nabla-2}
\end{equation}
the Laplacian of some function $\phi\in C^2$, by $\Omega^\circ$ we denote the interior of the domain $\Omega$, and by $\partial\Omega$ its piecewise smooth boundary; here $u_0$ is some  initial heat distribution, $\theta_0$ is the constant ambient space temperature ($20^\circ C$), and we use the shorthands 
%--
%--
\begin{equation}
\nonumber
\alpha\coloneqq \frac{\kappa}{c \rho}
%\label{eq:alpha}
%\end{equation}
%--
\quad \textrm{ and } \quad
%--
%\begin{equation}
%\nonumber
\beta\coloneqq \frac{1}{c \rho}
\label{eq:alpha}
\end{equation}
%--
with $\kappa$, $c$ and $\rho$ all in $\mathbb{R}^{+}$, 
denoting thermal conductivity, specific heat capacity and mass density. 
Our power density distribution of choice to simulate a laser beam is a Gaussian function$\colon$
%--
\begin{equation}
 I\left(x,y\right) = I_0 \cdot \exp{ \left[-2 \left[ \left(\frac{x-x_c}{\omega}\right)^2 + \left(\frac{y-y_c}{\omega}\right)^2 \right] \right]}
 \label{eq:Heat}
\end{equation}
%--
with an intensity constant
%--
\begin{equation}
 I_0=  \frac{2 P}{\pi\omega^2}
 \label{eq:Heat-I}
\end{equation}
%--
by knowing $\omega$ and $P$ to be the radius of the Gaussian beam waist and the laser power, respectively. In our study we let $x\in\left[-1,+1\right]$, $y\in\left[-1,+1\right]$ with $\left(x,y\right)\in\Omega^\circ$, $t\in\mathbb{R}^{+}$ and also $u\left(x,y,0\right)=0$. The aluminium thermal properties are used to simulate the metal powder spread across the manufacturing plate. 

We solved~\eqref{eq:HE} using~\cite{Matlab} by setting $P=4200\left(\si{\watt}\right)$ and $\omega=35$ pixels while letting $\left(x_c,y_c\right)$ to take all possible trajectory points such that the domain $\Omega^\circ$ always be affected by five consecutive heat source moves. In this way, we simulated the heat source movement across the board.\\

\textbf{Control objective.}
By adoption of the TSP based protocol, we aim to minimize the value of a desired objective function$\colon$
\begin{equation}
 J(m) = \frac{1}{2} \int_{0}^{T}
                               \left(
                               \int_{\Omega}
                               |\nabla u_m(z,t) |^2 +  
                               \left(u_m(z,t)-u_{g}\right)^2
                               \dd{z}
                               \right)
                               \dd{t}
 \label{eq:cost-F}
\end{equation}
%--
with 
\begin{equation}
    u_m=u_m(z,t)\,,\quad z=(x,y)\in\Omega^\circ, \, t\in[0,T]
\end{equation}
 being the solution of the heat equation (\ref{eq:HE}) on the interval $[0,T]$ under the control 
 \begin{equation}
     m\colon[0,T]\to\Omega^\circ\,, \quad t\mapsto(x_c(t), y_c(t))
 \end{equation}
that describes the trajectory of the center of the laser beam. Moreover, we have introduced $u_{g}$ as the desired target temperature to be maintained over the domain $\Omega^\circ$ as time $t$ evolves.

The motivation behind~\eqref{eq:cost-F} is to maintain a smooth temperature gradient over the entire plate for all $t\in[0,T]$ as is achieved by minimizing the $L^2$-norm of the gradient, $\nabla u$, while at the same time keeping the (average) temperature near a desired temperature $u_g$ for any time $t$.

\begin{figure} 
\centering
\includegraphics[width=.44\textwidth]{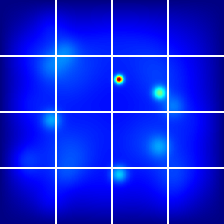}
\caption{We divide the entire domain $\Omega^\circ$ containing the diffused temperature values into $4\times 4$ sub-domains separated by white lines. Within each sub-domain,~\eqref{eq:cost-F} is computed to reveal how the temperature gradient $|\nabla{u}_m(\cdot,t)|$ evolves as a function of time $t$ evolves and how the average temperature $\bar{u}$ is maintained near to a target value of $u_g$. Note, the laser beam positions are in this image irrelevant.}
\label{fig:localGloabl} 
\end{figure}

We proceed, by dividing the entire plate $\Omega^\circ$ into $4\times 4$ sub-domains (see Fig.~\ref{fig:localGloabl}) and investigate our objective function~\eqref{eq:cost-F} within each sub-domain as explained in Section~\ref{sec:Results}. 
%-------------
%-----------
%%%%%%%%%%%%%%%%%%%%%%%%%%%%%%%%%%
\subsection{TSP-based Formulation}
\label{sec:TSP}
%%%%%%%%%%%%%%%%%%%%%%%%%%%%%%%%%%
A common assumption among numerous variants of the TSP~\cite{F1956}, as an NP-hard problem~\cite{GJ1990}, is that a set of cities have to be visited on a shortest possible \textit{tour}. Let $\mathcal{C}_{n\times n}$ be a symmetric matrix specifying the distances corresponding to the paths connecting a vertex set $\mathcal{V}=\left\{1,2,3,\cdots,n\right\}$ of cities to each other with $n\in\mathbb{N}$ to be the number of cities. A tour over the complete undirected graph $\mathcal{G}\left(\mathcal{V},\mathcal{C}\right)$ is defined as a cycle passing through each vertex exactly once. The traveling salesman problem seeks a tour of minimum distance.  

To adopt the TSP into our context, we formulate its input $\mathcal{V}$ as the set of all $16\times 16$ stopping points of the heat source over the board $\Omega^\circ$, and the set $\mathcal{C}$ as a penalty matrix with each element $\mathcal{C}_{ij}\ge 0$ being the impact (i.e. cost) of moving the heat source from a $i\in\mathcal{V}$ to $j\in\mathcal{V}$. For every vertex $i\in\mathcal{V}$, the possible movements to all $j\neq i$ with the associated cost \eqref{eq:cost-F} is computed and assigned to $\mathcal{C}_{ij}$ (see below for details). With this formulation, we want to remind the reader that $\mathcal{C}$ elements are nonnegative and follow the triangle inequality$\colon$
\begin{equation}
\mathcal{C}_{ij}\le\mathcal{C}_{ik}+\mathcal{C}_{kj}
\label{eq:TriIne}
\end{equation}
with $i,j,k\in\mathcal{V}$.

Note that, $\mathcal{C}$ matrix is obtained based on a prior set of temperature maps produced using FEM without enforcing any particular protocol on them. 

With this general formulation at hand, let us have a closer look at the discretized form of~\eqref{eq:cost-F} that was used in current study to compute the elements of the penalty matrix: 
\begin{equation}
                \small
                \mathcal{C}_{ij}=
                            \abs{   
                              \sum_{l=1}^{4\times 4}{ 
                              \big(
                               \left\|\boldsymbol{\Psi}\left(i,l\right) \right\|^2
                               +  
                               \left(\boldsymbol{\Lambda}\left(i,l\right)-u_{g}\right)^2 
                               \big)
                               }
                               - 
                              \sum_{l=1}^{4\times 4}{                                
                               \big(
                                \left\|\boldsymbol{\Psi}\left(j,l\right) \right\|^2
                               +  
                               \left(\boldsymbol{\Lambda}\left(j,l\right)-u_{g}\right)^2                                
                               \big)
                               }
                              }
\label{eq:discrCost}
\end{equation}
with $l$ to be the sub-domain index. In addition, $\small\boldsymbol{\Psi}\left(\cdot,l\right)=\sum_{z\in\Omega_l}\sum_{t\in t_l}{\nabla{u}_m\left(z,t\right)}$ represents the temperature gradient aggregation within each sub-domain, and $\small\boldsymbol{\Lambda}\left(\cdot,l\right)=\frac{1}{\abs{\Omega_l}}\sum_{z\in\Omega_l}\sum_{t\in t_l}{u_m\left(z,t\right)}$ is the average temperature value of each sub-domain, with $t_l$ to be the time period on which the nozzle operates on $\Omega_l$. Here, by $\abs{\Omega_l}$, we mean the number of discrete points in $\Omega_l\subset\Omega^\circ$. In other words, \eqref{eq:discrCost} is the TSP cost of moving the nozzle from the $i^{\text{th}}$ to the $j^{\text{th}}$ stopping point that depends on (a) the mean square deviation of the temperature field from constancy and (b) on the mean square deviation from the global target temperature $u_g$.  
In our simulation, the nozzle moves in the direction of the shortest (Euclidean) path connecting two successive stopping points. Thereby we assume the nozzle always adjusts its velocity so that the path between any arbitrarily chosen stopping points $i$ and $j$ always takes the same amount of time. The motivation behind this is to avoid heating up the entire domain $\Omega^\circ$ as a result of keeping the nozzle velocity constant. %Overheating can occur for those successive stopping points far from each other that are frequently visited by the laser. % adopting TSP.
 
In practice no polynomial-time algorithm is known for solving the TSP~\cite{GJ1990}, so we adopt a~\textit{simulated annealing algorithm}~\cite{TMZ1999} that was first proposed in statistical physics as a means of determining the properties of metallic alloys at a given temperatures \cite{MRRTT1953}. In the TSP context, we adopt~\cite{TMZ1999} to look for a good (but in general, sub-optimal) tour corresponding to the movement of the heat source leading to the minimization of~\eqref{eq:discrCost}.

\begin{figure} 
    \centering
    \begin{subfigure}[b]{.32\textwidth}
        \includegraphics[width=\textwidth]{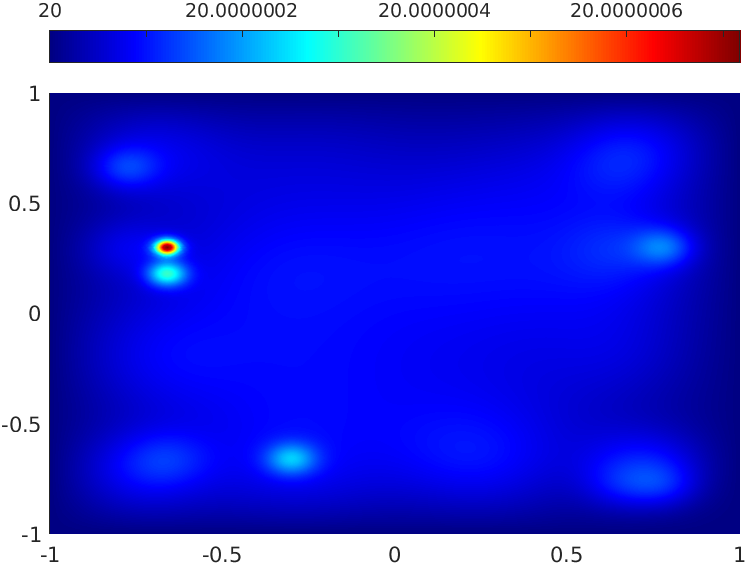}
    \end{subfigure}
    \begin{subfigure}[b]{.32\textwidth}
        \includegraphics[width=\textwidth]{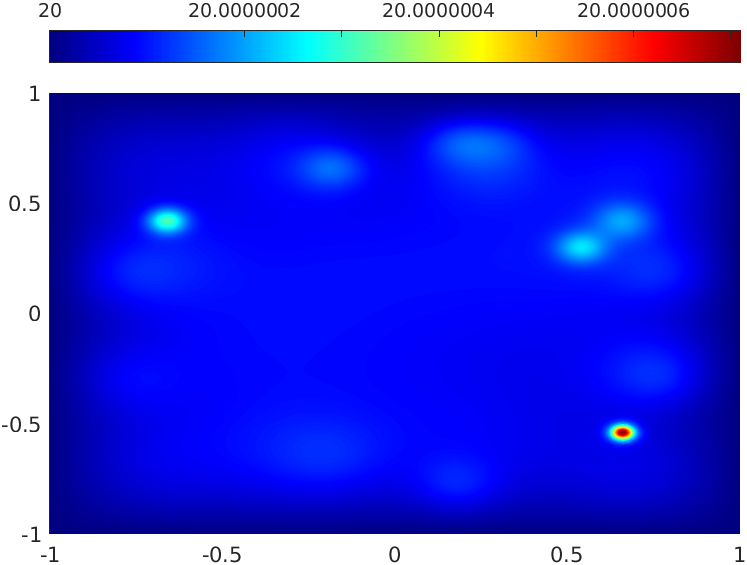}
    \end{subfigure}
    \begin{subfigure}[b]{.32\textwidth}
        \includegraphics[width=\textwidth]{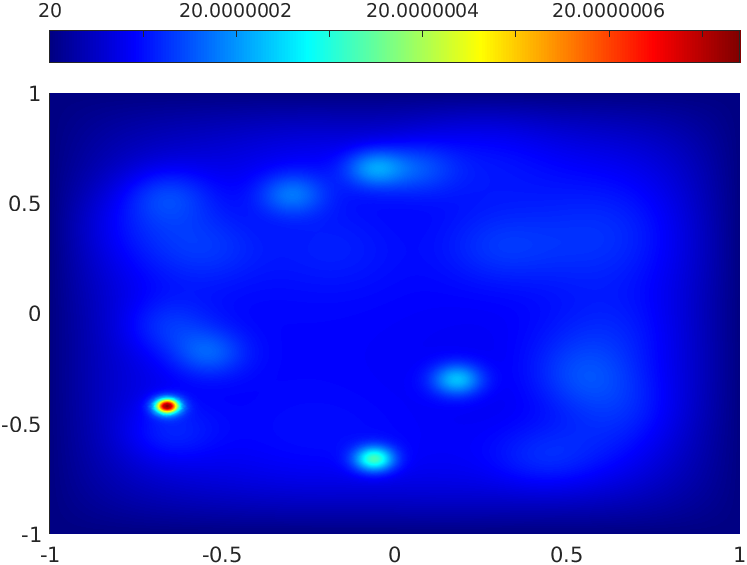}
    \end{subfigure}
    \begin{subfigure}[b]{.32\textwidth}
        \includegraphics[width=\textwidth]{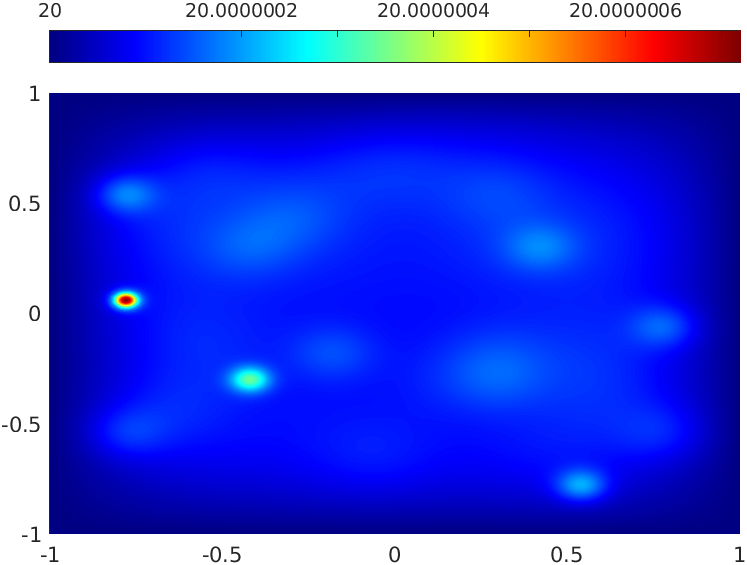}
    \end{subfigure}    
    \begin{subfigure}[b]{.32\textwidth}
        \includegraphics[width=\textwidth]{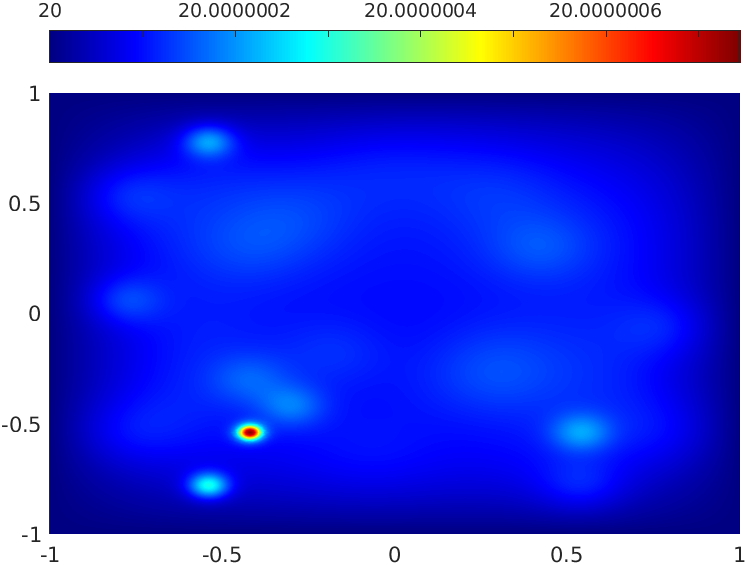}
    \end{subfigure}
    \begin{subfigure}[b]{.32\textwidth}
        \includegraphics[width=\textwidth]{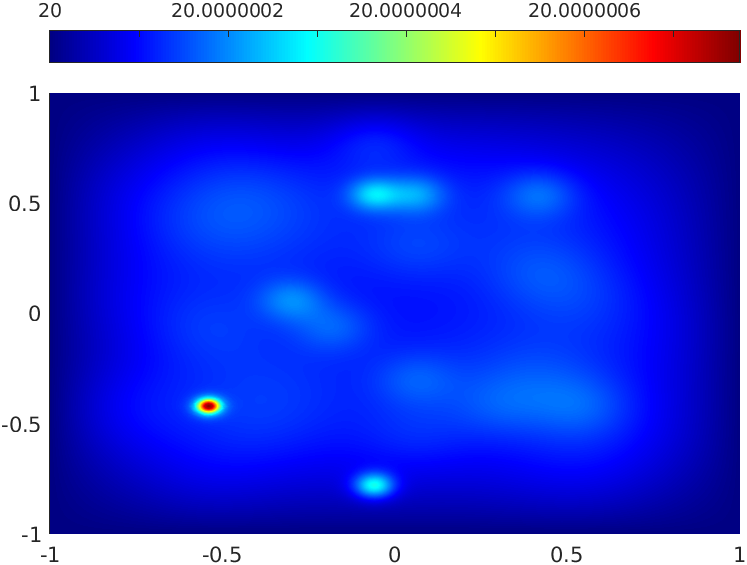}
    \end{subfigure}
    \caption{A subset of heat maps produced by FEM as the solution to the heat equation~\eqref{eq:HE}. One clearly observes the effect of the previous laser positions on current status of the map, in terms of diffused temperature. The TSP as a heuristics steers the heat source across the plate aiming to keep temperature constancy. Note that all temperatures are in Celsius.}
    \label{fig:TSPMaps}
\end{figure}

In Section~\ref{sec:Results}, we reveal our prediction results obtained by adopting a TSP based heuristic along with the LSTM network. Before moving to the next section, let us observe a subset of temperature maps obtained based on TSP shown as Fig.~\ref{fig:TSPMaps}.

%%%%%%%%%%%%%%%%%%%%%%%%%%%%%%
\section{The LSTM Approach}
\label{sec:LSTM}
\begin{figure}[htb]
\begin{center}    
   \begin{minipage}[b]{.55\linewidth}
   \includegraphics[width=\textwidth]{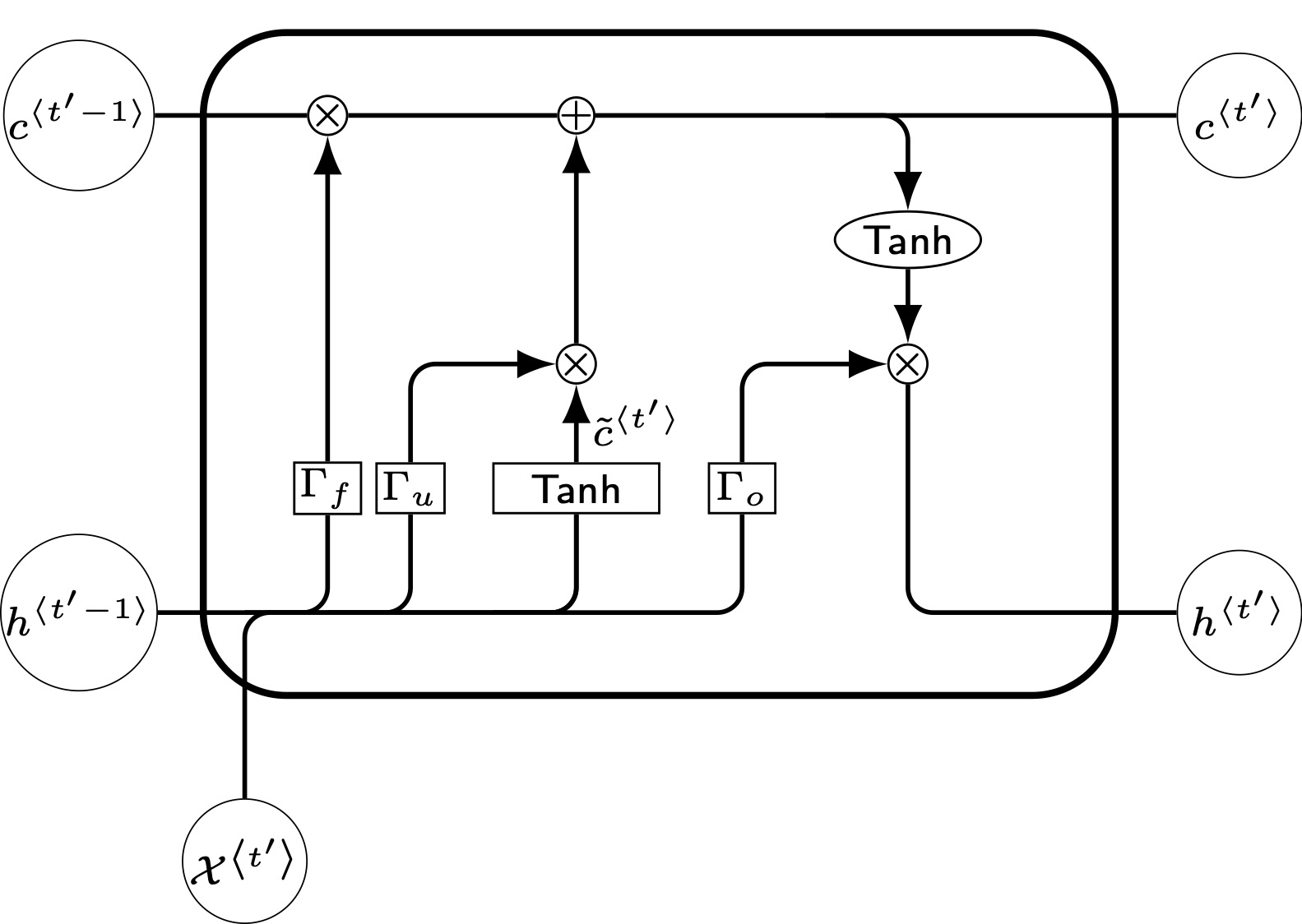}
   \subcaption{}
   \label{fig:LSTMsLEFT}
   \end{minipage}
   \begin{minipage}[b]{.41\linewidth}
   \includegraphics[width=.8\textwidth]{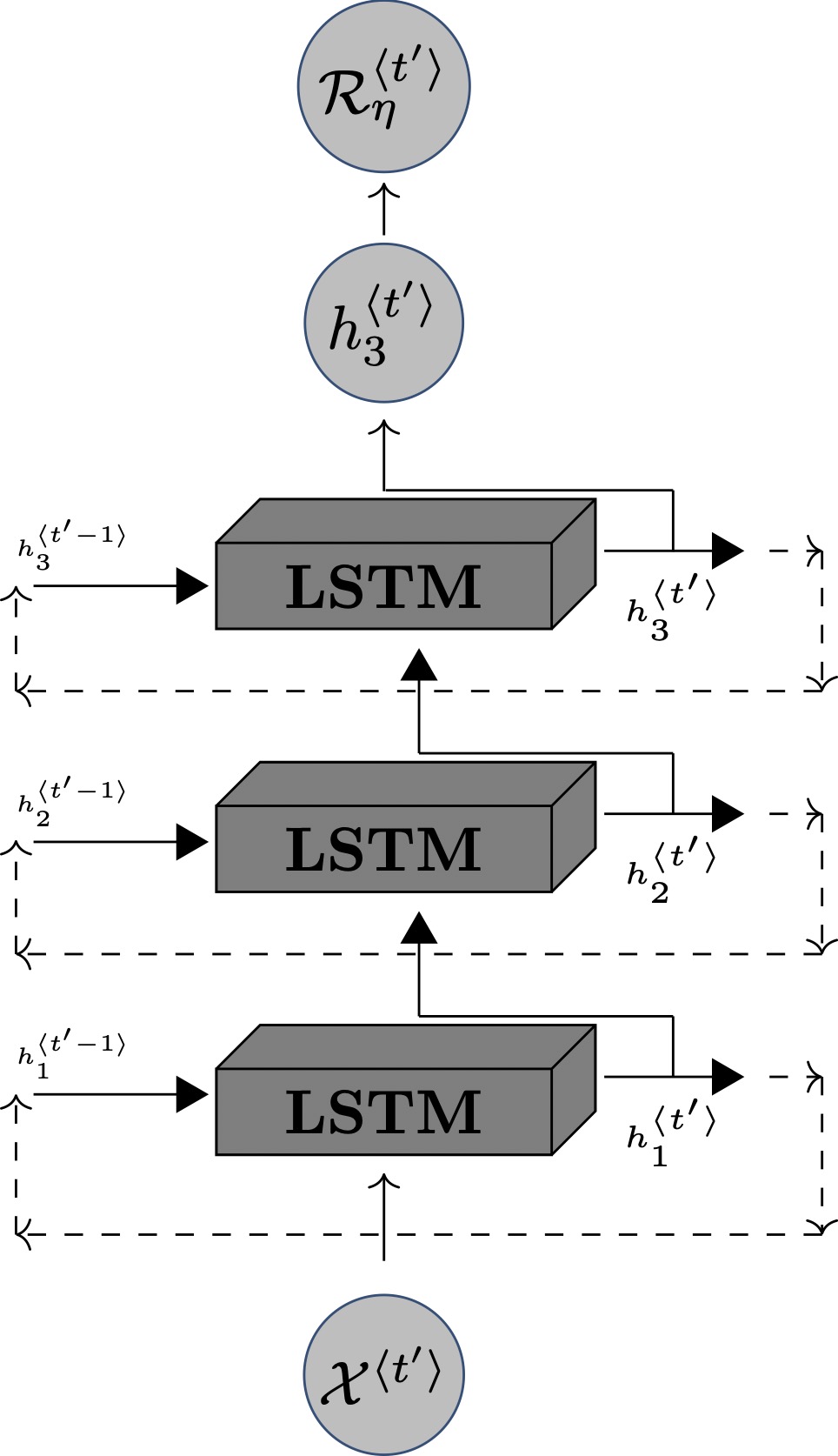} 
   \subcaption{}
   \label{fig:LSTMsRIGHT}   
   \end{minipage}
\end{center}
\caption{(a) A graphical representation of the LSTM cell accepting the hidden state $h^{\langle t^\prime-1\rangle}$ and the cell state $c^{\langle t^\prime-1\rangle}$ from the previous LSTM and the feature vector $\mathcal{X}^{\langle t^\prime\rangle}$ at the current time. (b) A schematic representation of the adopted stack of LSTMs comprised of three recurrent layers processing the data. The upper LSTM layer is followed by a fully connected layer. The network performs a regression task being trained based on half-mean-square-error loss function~\eqref{eq:hmse}. Note, the fully connected layer is established between the output of LSTM stack $h_3^{\langle t^\prime\rangle}$ and the only neuron of the last layer $\mathcal{R}^{\langle t^\prime\rangle}_\eta$ with $\eta=1$.}
\label{fig:LSTM} 
\end{figure}

%%%%%%%%%%%%%%%%%%%%%%%%%%%%%%
%
% 
%
Let us start discussion of our deep learning framework structure by investigating its LSTM~\cite{HS1997} cell building blocks shown as Fig.~\ref{fig:LSTMsLEFT} used to comprise a stack of three LSTM layers (see Fig.~\ref{fig:LSTMsRIGHT}) followed by a fully connected layer.

Here, we use temperature gradient values of $\mu=14$ previous (i.e.\ from previous time) heat maps to predict the gradients values of the current heat map. By letting $\zeta$ to be the current heat map, its history feature values formally lie in a range of $[\zeta-\mu,\zeta-1]$ heat maps with $\zeta>\mu$. By considering each heat map to have $16$ sub-domains and the same number of gradient features $\Psi\left(\cdot,l\right)$, each corresponding to one sub-domain, we obtain in total $\nu=\mu\times 16$ number of gradient feature history values that we vectorise to frame the vector $\mathcal{X}\in\mathbb{R}^{\nu}$. Our aim is to use sub-sequences from $\mathcal{X}$ to train the stacked of LSTMs and forecast a sequence of $16$ number of gradient feature values corresponding to the sub-domains of a heat map of interest $\zeta$. %Let us note that, each LSTM cell has $q\in\mathbb{N}$ number of hidden units.
 
Let us briefly discuss the weight and bias matrix dimensions of each LSTM cell. Here, we use $q\in\mathbb{N}$ as the number of hidden units of each LSTM cell and $n\in\mathbb{N}$ to represent the number of features that we obtain from FEM based heat maps and fed to the LSTM cell. More specifically, we have only one feature $\Psi\left(\cdot,l\right)$ per sub-domain, i.e.\ $n=1$. In practice, during the training process and at a particular time $t^\prime$, a batch of input feature values $\mathcal{X}\supset\mathcal{X}^{\langle t^\prime\rangle}\in\mathbb{R}^{b\times n}$ with $b\in\mathbb{N}$ to be the batch size, are fed to each LSTM cells of the lowest stack level in Fig.~\ref{fig:LSTMsRIGHT}. Here LSTM learns to map each feature value in $\mathcal{X}^{\langle t^\prime\rangle}$ to its next adjacent value in $\mathcal{X}$ as its label. The mapping labels are applied during the training and to the only neuron $\mathcal{R}_\eta\in\mathbb{R}$ of the last fully connected layer with $\eta=1$.

In addition to $\mathcal{X}^{\langle t^\prime\rangle}$, each LSTM cell accepts two others inputs, namely $h^{\langle t^\prime-1\rangle}\in\mathbb{R}^{b\times q}$ and $c^{\langle t^\prime-1\rangle}\in\mathbb{R}^{b\times q}$, the so called the \textit{hidden state} and \textit{cell state} both of which are already computed at the time $t^\prime-1$. Here, the cell state $c^{\langle t^\prime-1\rangle}$ carries information from the intervals prior to $t^\prime$. 

%-------------------
%-------------------
A few remarks on how the cell state $c^{\langle t^\prime\rangle}$~\eqref{eq:MemoryCell} at time $t^\prime$ is computed by the formula~\eqref{eq:MemoryCell} below are in order: A more precise look at~\eqref{eq:MemoryCell} reveals that it partially depends on $\Gamma_f\odot c^{\langle t^\prime-1\rangle}$ with $c^{\langle t^\prime-1\rangle}$ being the cell state at the previous time $t^\prime-1$. The term $\Gamma_f$ satisfies 
%---------------
\begin{equation}
c^{\langle t^\prime\rangle}=\Gamma_u\odot\tilde{c}^{\langle t^\prime\rangle}+\Gamma_f\odot c^{\langle t^\prime-1\rangle}
\label{eq:MemoryCell}
\end{equation}
%---------------
with $\odot$ representing the element-wise vector multiplication. As~\eqref{eq:MemoryCell} further shows, $c^{\langle t^\prime\rangle}$ also depends on $\tilde{c}^{\langle t^\prime\rangle}$ that itself is computed based on the feature vector $\mathcal{X}^{\langle t^\prime\rangle}$ and the previous hidden state $h^{\langle t^\prime-1\rangle}$ as$\colon$
%---------------
\begin{equation}
\tilde{c}^{\langle t^\prime\rangle}=\tanh{\left(\left[\left(h^{\langle t^\prime-1\rangle}\right)_{b\times q}\vline\hspace{.5mm}\left(\mathcal{X}^{\langle t^\prime\rangle}\right)_{b\times n}\right]\times\mathcal{W}_{c}+\left(b_c\right)_{b\times q}\right)}
\label{eq:C-Hat}
\end{equation}
%---------------
with $\times$ visualizing in this work standard matrix multiplication, and $b_c$ and $\mathcal{W}_{c}$ to be the corresponding bias and weight matrices, respectively.

Equation~\eqref{eq:MemoryCell} contains two further terms, $\Gamma_u$ and $\Gamma_f$, called the \textit{update gate} and \textit{forget gate} defined as
%---------------
\begin{equation}
\Gamma_u=\sigma{\left(\left[\left(h^{\langle t^\prime-1\rangle}\right)_{b\times q}\vline\hspace{.5mm}\left(\mathcal{X}^{\langle t^\prime\rangle}\right)_{b\times n}\right]\times \mathcal{W}_{u}+\left(b_u\right)_{b\times q}\right)}
\label{eq:UpdateGate}
\end{equation}
%---------------
and
%---------------
\begin{equation}
\Gamma_f=\sigma{\left(\left[\left(h^{\langle t^\prime-1\rangle}\right)_{b\times q}\vline\hspace{.5mm}\left(\mathcal{X}^{\langle t^\prime\rangle}\right)_{b\times n}\right]\times \mathcal{W}_{f}+\left(b_f\right)_{b\times q}\right)}
\label{eq:ForgetGate}
\end{equation}
%---------------
that are again based on the feature vector $\mathcal{X}^{\langle t^\prime\rangle}$ and the previous hidden state $h^{\langle t^\prime-1\rangle}$ with $b_u$, $b_f$, $\mathcal{W}_{u}$ and $\mathcal{W}_{f}$ to be the corresponding biases and weight matrices. Let us shortly conclude here, that the feature vector $\mathcal{X}^{\langle t^\prime\rangle}$ and the previous hidden state $h^{\langle t^\prime-1\rangle}$ are the essential ingredients used to compute $\tilde{c}^{\langle t^\prime\rangle}$, $\Gamma_u$ and $\Gamma_f$, that are all used to update the current cell state $c^{\langle t^\prime\rangle}$~\eqref{eq:MemoryCell}.

The motivation to use the \textit{sigmoid function} $\sigma$ in structure of the gates shown in~\eqref{eq:UpdateGate} and~\eqref{eq:ForgetGate} is its activated range in $\left[0,1\right]$, leading them in extreme cases to be fully on or off letting all or nothing to pass through them. In non-extreme cases they partially contribute the previous cell state $c^{\langle t^\prime-1\rangle}$ and the on the fly computed value $\tilde{c}^{\langle t^\prime\rangle}$ to the current cell state $c^{\langle t^\prime\rangle}$ as shown in~\eqref{eq:MemoryCell}. 

To give a bigger picture, let us visualize the role of the $\Gamma_u$ and $\Gamma_f$ gates concerning the cell state $c^{\langle t^\prime\rangle}$. In Fig.~\ref{fig:LSTMsLEFT}, a direct line connecting $c^{\langle t^\prime-1\rangle}$ to $c^{\langle t^\prime\rangle}$ carries the old data directly from time $t^\prime-1\rightarrow t^\prime$. Here, one clearly observes the $\Gamma_u$ and $\Gamma_f$ gates both are connected by $+$ and $\times$ operators to the passed line. They linearly contribute, as shown in~\eqref{eq:UpdateGate} and~\eqref{eq:ForgetGate}, the current feature value $\mathcal{X}^{\langle t^\prime\rangle}$ and the adjacent hidden state $h^{\langle t^\prime-1\rangle}$ to update the current cell state $c^{\langle t^\prime\rangle}$. Meanwhile, $\Gamma_u$ shares its contribution through $\times$ operator with $\tilde{c}^{\langle t^\prime\rangle}$ to the passing line.

Finally, to make the current LSTM activated we need the cell state value at the time $t^\prime$, namely $c^{\langle t^\prime\rangle}$, that we obtain from~\eqref{eq:MemoryCell} and also the so called \textit{output gate} obtained from
%---------------
\begin{equation}
\Gamma_o=\sigma{\left(\left[\left(h^{\langle t^\prime-1\rangle}\right)_{b\times q}\vline\hspace{.5mm}\left(\mathcal{X}^{\langle t^\prime\rangle}\right)_{b\times n}\right]\times\mathcal{W}_{o}+\left(b_o\right)_{b\times q}\right)}
\label{eq:OutPutGate}
\end{equation}
%---------------
with $\Gamma_o\in\left[0,1\right]$, and $b_o$ and $\mathcal{W}_{o}$ to be the corresponding bias and weight matrices. The final activated value of the LSTM cell is computed by
%---------------
\begin{equation}
h^{\langle t^\prime\rangle}=\Gamma_o\odot\tanh{\left(c^{\langle t^\prime\rangle}\right)}.
\label{eq:Activated}
\end{equation}
%---------------
Here, the obtained activated value $h^{\langle t^\prime\rangle}$ from~\eqref{eq:Activated} will be used as the input hidden state to the next LSTM cell at the time $t^\prime+1$. 

Let us also mention that all the biases $b_c,b_u,b_f,b_o\in\mathbb{R}^{b\times q}$ and the weight matrices are further defined as
%---------------
\begin{align*}
  \mathcal{W}_c \coloneqq \left[\begin{array}{cc}
        \left(\mathcal{W}_{ch}\right)_{q\times q} \vline\left(\mathcal{W}_{cx}\right)_{n\times q}  
    \end{array}\right]^\top, 
    \hspace{15mm}
    \mathcal{W}_u \coloneqq \left[\begin{array}{cc}
        \left(\mathcal{W}_{uh}\right)_{q\times q} \vline\left(\mathcal{W}_{ux}\right)_{n\times q}  
    \end{array}\right]^\top
\end{align*}
%---------------
%---------------
\begin{align*}
  \mathcal{W}_f \coloneqq \left[\begin{array}{cc}
        \left(\mathcal{W}_{fh}\right)_{q\times q} \vline\left(\mathcal{W}_{fx}\right)_{n\times q}  
    \end{array}\right]^\top,  
    \hspace{15mm}
  \mathcal{W}_o \coloneqq \left[\begin{array}{cc}
        \left(\mathcal{W}_{oh}\right)_{q\times q} \vline\left(\mathcal{W}_{ox}\right)_{n\times q}  
    \end{array}\right]^\top      
\end{align*}
%---------------
leading both the $\tilde{c}^{\langle t^\prime\rangle},c^{\langle t^\prime\rangle}\in\mathbb{R}^{b\times q}$.

Finally, we have a fully connected layer that maps the output $h_{b\times q}^{<\cdot>}$ of the stacked LSTM to the only neuron of the output layer $\mathcal{R}_\eta$. This is achieved during the training process while the weight matrix $\hat{\mathcal{W}}\in\mathbb{R}^{q\times\eta}$ and bias vector $\hat{b}\in\mathbb{R}^{b\times \eta}$ corresponding to the fully connected layer are updated based on a computed loss $\mathcal{L}$ using half-mean-square-error~\eqref{eq:hmse} between the network predictions and target temperature gradient values obtained from the heat maps produced by FEM.
\begin{equation}
\mathcal{L}=\frac{1}{2\eta b}\sum_{i_1=1}^{b}\sum_{i_2=1}^{\eta}\left(p_{i_1i_2}-y_{i_1i_2}\right)^2
\label{eq:hmse}
\end{equation}
Here, $p$ and $y$ values represent the predicted and the target gradient temperature values, respectively.
%
%
%%%%%%%%%%%%%%%%%%%%
\section{Results}
\label{sec:Results}
\begin{figure} 
\centering
\includegraphics[width=.55\textwidth]{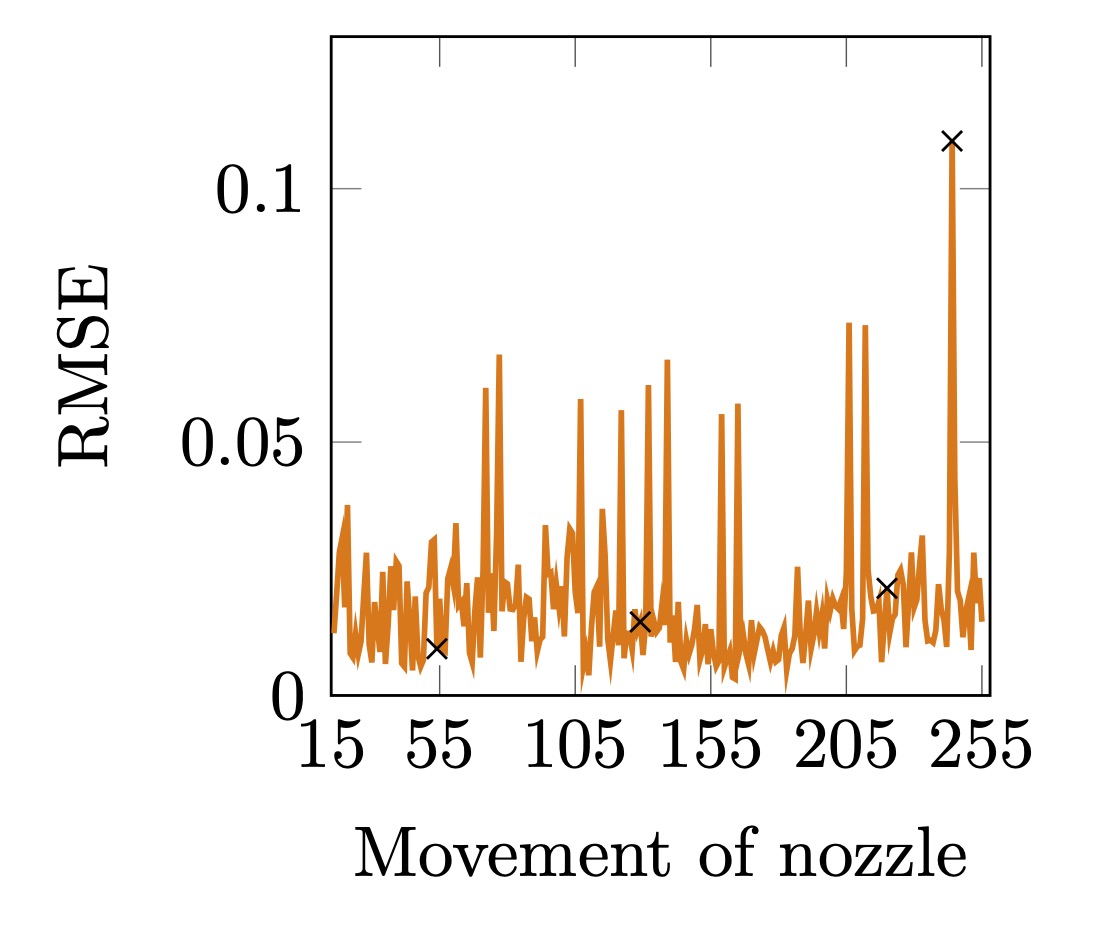}
\caption{Each curve value represents a computed RMSE between all $4\times 4$ sub-domains gradient feature values of predicted heat map $\zeta$ and their ground truth counterpart. The RMSE computation can be started from $15^\text{th}$ nozzle move onward, since we use a history of $\mu=14$ previous gradient maps. Those RMSE measures, highlighted as $\times$ in ascending order correspond to the $25^\text{th}$, $50^\text{th}$, $75^\text{th}$ and $100^\text{th}$ percentiles, respectively.} 
\label{fig:RMSE-Curve}
\end{figure}

%%%%%%%%%%%%%%%%%%%%
To begin with, we consider a set of computed root-mean-square-error measures (RMSE) between the predicted and the target gradient values corresponding to the nozzle moves as shown in Fig.~\ref{fig:RMSE-Curve}. More precisely, each curve value represents a computed RMSE between all $4\times 4$ sub-domains gradient feature values of predicted heat map $\zeta$ and their ground truth counterpart. Since we use a history of $\mu=14$ previous gradient heat maps, the first prediction can be performed for the $15^\text{th}$ nozzle move. Among all the measured RMSE values, we highlight four of them as can be seen in Fig.~\ref{fig:RMSE-Curve}, that correspond to the $25^\text{th}$, $50^\text{th}$, $75^\text{th}$ and $100^\text{th}$ percentiles. 
%--------------------------------------
%--------------------------------------

\begin{figure}
    \begin{subfigure}[h]{\textwidth}
        \includegraphics[width=.95\textwidth]{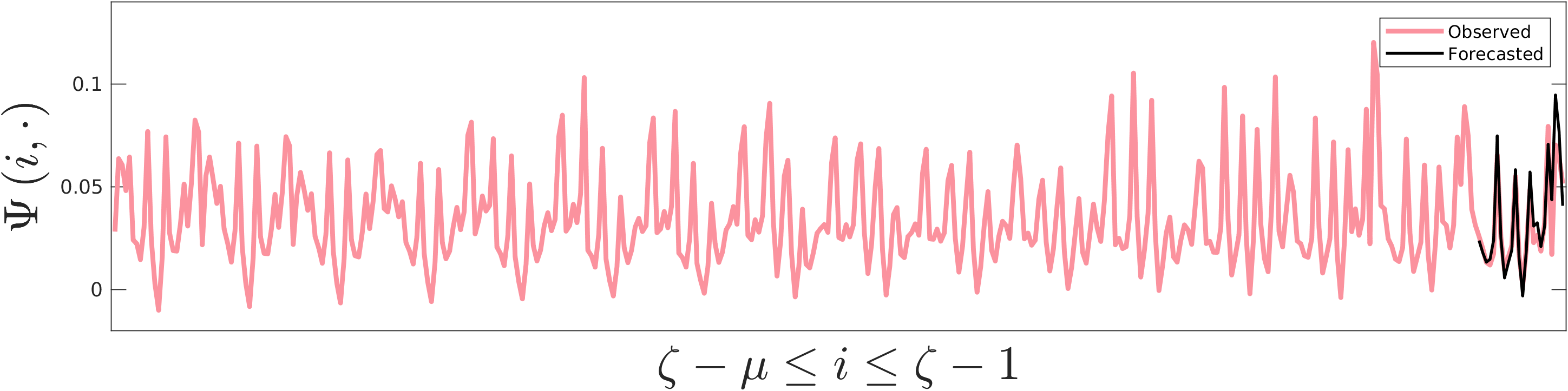}
        \caption{RMSE=$0.009$}
        \label{fig:PErcentiles-a}
    \end{subfigure}
    \begin{subfigure}[h]{\textwidth}
        \includegraphics[width=.95\textwidth]{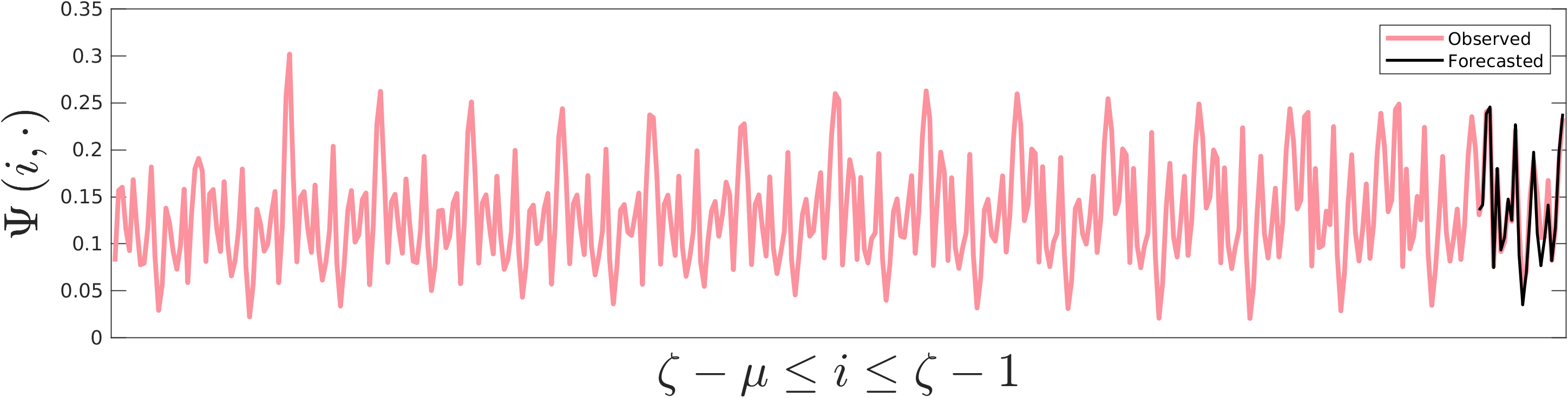}
        \caption{RMSE=$0.014$}
        \label{fig:PErcentiles-b}
    \end{subfigure}
    \begin{subfigure}[h]{\textwidth}
        \includegraphics[width=.95\textwidth]{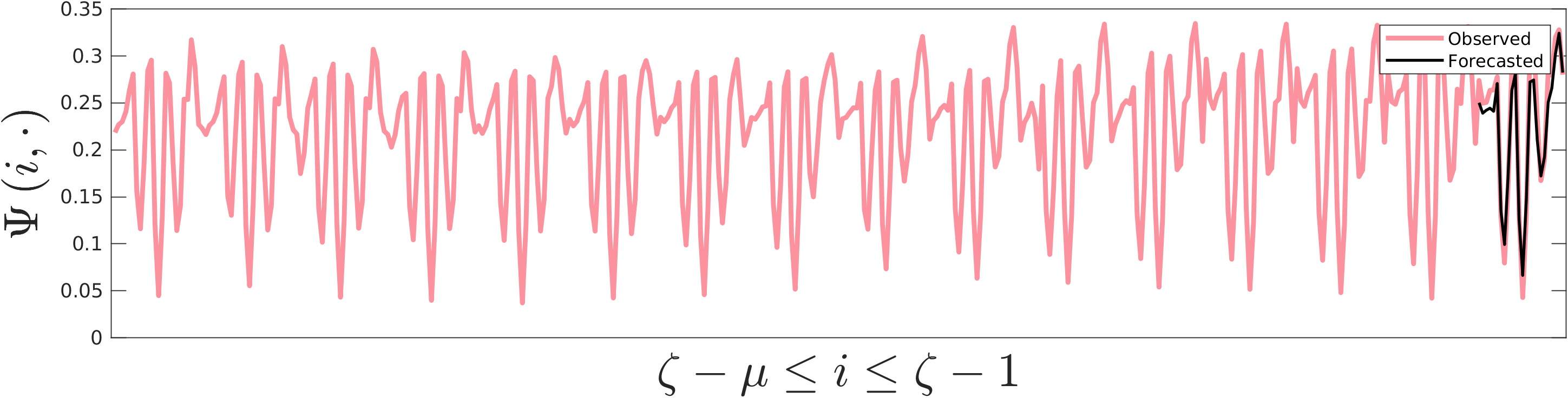}
        \caption{RMSE=$0.021$}
        \label{fig:PErcentiles-c}
    \end{subfigure}
    \begin{subfigure}[h]{\textwidth}
        \includegraphics[width=.95\textwidth]{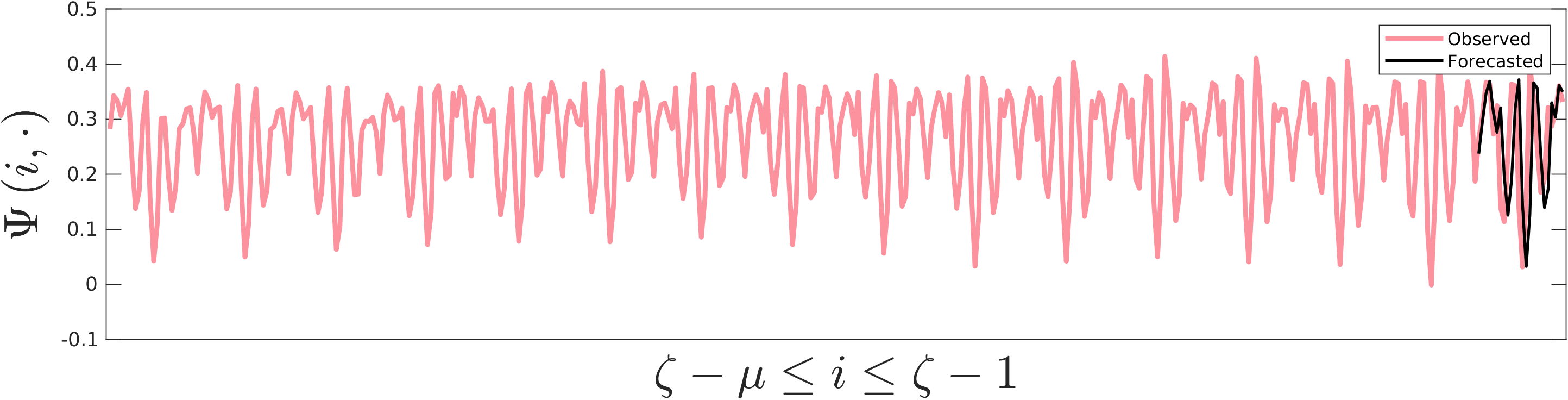}
        \caption{RMSE=$0.109$}
        \label{fig:PErcentiles-d}
    \end{subfigure}    
    \caption{The ground truth and predicted gradient feature values used to compute the RMSE measures of $25^\text{th}$, $50^\text{th}$, $75^\text{th}$ and $100^\text{th}$ percentiles shown in (a), (b), (c) and (d), respectively. The ground truth pink curves obtained and vectorized from $\mu$ number of previous heat maps preceding to the current heat map of $\zeta$. Here we use $i$ as the nozzle move number to vary in its range $\left[\zeta-\mu,\zeta-1\right]$ producing a set of gradient feature history values shown in pink. In addition the vectorized ground truth gradient feature values of the current heat map $\zeta$ are also shown as part of the pink curve that overlaps with the black curve. The black curve is the forecasted vectorized gradient feature values corresponding to the $\zeta$ heat map. We also have noted the RMSE measure concerning each case (a) to (d) below it. The last case (d) with the worst RMSE (See Fig.~\ref{fig:RMSE-Curve}) shows a clear asynchronous prediction, though the shape of the forecasted curve looks to conform with its overlapping ground truth. In cases (a), (b) and (c) we have synchronous predictions though in some parts the black curve is not predicting the pink one accurately. }
    \label{fig:PErcentiles}    
\end{figure}

As one observes in Fig.~\ref{fig:RMSE-Curve}, a relatively low RMSE measure is obtained almost across all nozzle moves on horizontal axis, though there exist some outliers. We further visualize the corresponding prediction results of the percentiles as Fig.~\ref{fig:PErcentiles}. Specifically, let us take the $25^\text{th}$ RMSE percentile computed between the black and its overlapping part of the pink curve shown in Fig.~\ref{fig:PErcentiles-a}. The black curve in Fig.~\ref{fig:PErcentiles-a} is comprised of $4\times 4$ forecasted vectorized gradient feature values of heat map sub-domains produced by $54^\text{th}$ nozzle move with a RMSE equal to $0.009$, compared to its overlapping pink curve. In this case, we let the $i$ as the nozzle move number to range in $\left[\zeta-\mu,\zeta-1\right]$ to produce a history of feature gradient values corresponding to the heat map with $\zeta=54$. This consequently means, the non-overlapping part of the pink curve in Fig.~\ref{fig:PErcentiles-a} represent the vectorized history feature values of $40^\text{th}$ to $53^\text{th}$ heat maps that comprise $\mu=14$ number of preceding heat maps of $\zeta=54$, each of them with $4\times 4$ sub-domains. The black curves in Figs.~\ref{fig:PErcentiles-b},~\ref{fig:PErcentiles-c} and~\ref{fig:PErcentiles-d} are also comprised of the predicted gradient feature values of the heat maps $\zeta$ equal to $129$, $220$ and $244$, respectively, that are forecasted based on $\mu$ number of their previous heat maps.
%--------------------------------------
%--------------------------------------

A closer look at the four prediction samples shown in Figs.~\ref{fig:PErcentiles} reveals that the even the $100^\text{th}$ percentile, that marks some kind of outlier, is accurately predicted in that the shape of the black curve tracks the pink curve (ground truth). Concerning other three RMSE percentile values, the synchronicity among the black and its pink curves is preserved equally well though in some part we do not have a full overlap.

Finally, the parameters used during the training phase are revealed to be the Adam optimizer~\cite{KB2014} applied on batch data of size $6$. The epoch value is chosen to be $350$ that results to a meaningful reduction of RMSE and loss measures within each batch. The initial learning rate was also chosen to be $0.008$ with a drop factor of $0.99$ concerning each $12$ epochs. To avoid the overfitting phenomenon, the flow of data within the network structure is randomly adjusted by setting the LSTMs outputs with a probability of $0.25$ to zero~\cite{SHKSS2014}.
%%%%%%%%%%%%%%%%%%%%
\section{Conclusion}
\label{sec:Conclusion}
%%%%%%%%%%%%%%%%%%%%
%
We developed a novel and practical pipeline and mathematically justified its comprising components. Our proposed model consists of two major components, namely the simulation part of a laser power bed fusion setup based on FEM and an intelligent agent based on LSTM network that actively judges the simulation results based on a proposed cost function. The FEM simulation can be robustly applied before conducting expensive real-world printing scenarios so that the intelligent component of the pipeline can decide on early stopping of the printing process. The LSTM based network predicts the forthcoming temperature rate of change across the simulated power bed based on previously seen temperature history leading us to have a means of control to achieve a final optimal printing process as visualized by our obtained results.
%%%%%%%%%%%%%%%%%%%%
\section*{Acknowledgements}
%%%%%%%%%%%%%%%%%%%%
The current work was supported by the European Regional Development Fund, EFRE 85037495.
%###########################
%
% ---- Bibliography ----
%


\begin{thebibliography}{6}
%
\bibitem{A-et-al-2019}
Ali, M., Porter, D., K\"{o}mi, J., Eissa, M., Faramawy, H., Mattar, T.: Effect of cooling rate and composition on microstructure and mechanical properties of ultrahigh-strength steels. {\em Journal Of Iron And Steel Research International}. pp. 1-16 (2019)
%
\bibitem{ARNST2017}
Abdelrahman, M., Reutzel, E., Nassar, A., Starr, T.: Flaw detection in powder bed fusion using optical imaging. {\em Additive Manufacturing}. \textbf{15}, 1-11 (2017)
%
\bibitem{BSM2018}
Baturynska, I., Semeniuta, O., Martinsen, K. : Optimization of process parameters for powder bed fusion additive manufacturing by combination of machine learning and finite element method: A conceptual framework. {\em Procedia Cirp}. \textbf{67} pp. 227-232 (2018)
%
\bibitem{F1956}
Flood, M.: The Traveling-Salesman Problem. {\em Operations Research}. \textbf{4}, 61-75 (1956)
%
\bibitem{FB2007}
Fish, J., Belytschko, T.: A first course in finite elements. Wiley (2007)
%
\bibitem{G-et-al-2021}
Großwendt, F., R\"{o}ttger, A., Strauch, A., Chehreh, A., Uhlenwinkel, V., Fechte-Heinen, R., Walther, F., Weber, S., Theisen, W.: Additive manufacturing of a carbon-martensitic hot-work tool steel using a powder mixture – Microstructure, post-processing, mechanical properties. {\em Materials Science And Engineering: A}. \textbf{827} pp. 142038 (2021)
%
\bibitem{GJ1990}
Lewis, H.: A guide to the theory of NP-completeness. {\em The Journal Of Symbolic Logic}. \textbf{48}, 498-500 (1983)
%
\bibitem{HS1997}
Hochreiter, S., Schmidhuber, J.: Long short-term memory. {\em Neural Computation}. \textbf{9}, 1735-1780 (1997)
%
\bibitem{KB2014}
Kingma, D., Ba, J. : Adam: A method for stochastic optimization. {\em ArXiv Preprint ArXiv:1412.6980} (2014)
%
\bibitem{KSF2015}
Kanko, J., Sibley, A., Fraser, J.: In situ morphology-based defect detection of selective laser melting through inline coherent imaging. {\em Journal Of Materials Processing Technology}. \textbf{231} (2015)
%
\bibitem{KZZ2014}
Krauss, H., Zeugner, T., Zaeh, M.: Layerwise monitoring of the selective laser melting process by thermography. {\em Physics Procedia}. \textbf{56} pp. 64-71 (2014)
%
\bibitem{LB2015}
Lecun, Y., Bengio, Y.,: Convolutional Networks for Images, Speech and Time Series. {\em The Handbook Of Brain Theory And Neural Networks}. pp. 255-258 (1995)
%
\bibitem{M-et-al-2018}
Mozaffar, M., Paul, A., Al-Bahrani, R., Wolff, S., Choudhary, A., Agrawal, A., Ehmann, K., Cao, J.: Data-driven prediction of the high-dimensional thermal history in directed energy deposition processes via recurrent neural networks. {\em Manufacturing Letters}. \textbf{18} pp. 35-39 (2018)
%
\bibitem{Matlab-KNN}
The MathWorks, k-Means Clustering. (2020)
%
\bibitem{Matlab}
The MathWorks, Partial Differential Equation Toolbox.  (2020)
%
\bibitem{MRRTT1953}
Metropolis, N., Rosenbluth, A., Rosenbluth, M., Teller, A., Teller, E.: Equation of State Calculations by Fast Computing Machines . {\em J. Chem. Phys.}. \textbf{21} pp. 1087 (1953)
%
\bibitem{SHRUBTWZ2020}
Taruttis, A., Hardes, C., R\"{o}ttger, A., Uhlenwinkel, V., Chehreh, A., Theisen, W., Walther, F., Zoch, H., Laser additive manufacturing of hot work tool steel by means of a starting powder containing partly spherical pure elements and ferroalloys. {\em Procedia CIRP}. \textbf{94} pp. 46-51 (2020)
%
\bibitem{SCS2017}
Schoinochoritis, B., Chantzis, D., Salonitis, K.: Simulation of metallic powder bed additive manufacturing processes with the finite element method: A critical review. {\em Proceedings Of The Institution Of Mechanical Engineers, Part B: Journal Of Engineering Manufacture}. \textbf{231}, 96-117 (2017)
%
\bibitem{SB2018}
Scime, L., Beuth, J.: Anomaly Detection and Classification in a Laser Powder Bed Additive Manufacturing Process using a Trained Computer Vision Algorithm. {\em Additive Manufacturing}. \textbf{19} pp. 114-126 (2018)
%
\bibitem{SB2018-CNN}
Scime, L., Beuth, J.: A multi-scale convolutional neural network for autonomous anomaly detection and classification in a laser powder bed fusion additive manufacturing process. {\em Additive Manufacturing}. \textbf{24} pp. 273-286 (2018)
%
\bibitem{SFZSLMWYORK2020}
Song, X., Feih, S., Zhai, W., Sun, C., Li, F., Maiti, R., Wei, J., Yang, Y., Oancea, V., Romano Brandt, L., Korsunsky, A. : Advances in additive manufacturing process simulation: Residual stresses and distortion predictions in complex metallic components. {\em Materials and Design}. \textbf{193} pp. 108779 (2020)
%
\bibitem{SHKSS2014}
Srivastava, N., Hinton, G., Krizhevsky, A., Sutskever, I., Salakhutdinov, R. : Dropout: a simple way to prevent neural networks from overfitting. {\em The Journal Of Machine Learning Research}. \textbf{15}, 1929-1958 (2014)
%
\bibitem{TMZ1999}
Tian, P., Ma, J., Zhang, D.: Application of the simulated annealing algorithm to the combinatorial optimisation problem with permutation property: An investigation of generation mechanism.. {\em Eur. J. Oper. Res.}. \textbf{118}, 81-94 (1999)
%
\bibitem{ZC2006}
Zhang, Y., Chou, Y.: Three-dimensional finite element analysis simulations of the fused deposition modelling process. {\em Proceedings Of The Institution Of Mechanical Engineers, Part B: Journal Of Engineering Manufacture}. \textbf{220} pp. 1663 - 1671 (2006)
%
\bibitem{ZC2008-parametric}
Zhang, Y., Chou, K.: A parametric study of part distortions in fused deposition modelling using three-dimensional finite element analysis. {\em Proceedings Of The Institution Of Mechanical Engineers, Part B: Journal Of Engineering Manufacture}. \textbf{222} pp. 959 - 968 (2008)
%
\end{thebibliography}
\end{document}